\documentclass[letterpaper, 10 pt, conference]{ieeeconf}  

\IEEEoverridecommandlockouts                              

\usepackage{graphicx} 
\usepackage{epsfig} 
\usepackage{mathptmx} 
\usepackage{times} 
\usepackage{amsmath} 
\usepackage{amssymb} 
\usepackage{booktabs} 
\usepackage{multirow} 
\usepackage{algorithm} 
\usepackage{algpseudocode} 
\usepackage[hidelinks]{hyperref} 

\title{CABLE: Cloud-Assisted Bandwidth-efficient LMM-based Encoding \\
for V2X Systems}

\author{Haohua Que, Zhipeng Bao, Qianyi Wu, and Handong Yao%
\thanks{All authors are with the College of Engineering, University of Georgia, Athens, GA 30602, USA.
	{\tt\small Haohua.Que@uga.edu, Zhipeng.Bao@uga.edu, Qianyi.Wu@uga.edu, Handong.Yao@uga.edu}}%
}

\begin{document}

\maketitle
\thispagestyle{empty}
\pagestyle{empty}

\begin{abstract}
	Cloud-hosted large multimodal models (LMMs) can provide strong open-vocabulary perception for Vehicle-to-Everything systems, but naively transmitting full-resolution frames from edge to cloud causes severe communication overhead and high cloud-side prefill latency. We present CABLE, a cloud-assisted bandwidth-efficient LMM-based encoding framework for edge-cloud perception. CABLE propagates the previous cloud segmentation mask on the edge using ego-motion compensation, refines it with residual-motion cues, and consolidates disconnected regions via a corridor envelope to form a robust region of interest (ROI). Only ROI-masked images are uploaded, while the cloud segmentation output is fed back as the prior for the next frame, forming a mask-to-ROI-to-LMM feedback loop. Experiments on five datasets (nuScenes, WOD-ZB, Waymo, KITTI, and CADC) show consistent communication savings while largely preserving perception, achieving $73$--$87\%$ ROI pixel-coverage reduction with $5$--$8\times$ estimated LMM prefill speedup at a modest detection-quality trade-off relative to full-frame inference.

\end{abstract}

\section{Introduction}
\label{sec:intro}

Vehicle-to-Everything (V2X) communication enables cooperative perception by offloading compute-intensive tasks from resource-constrained edge devices to powerful cloud servers~\cite{jiang2023lidar,hawlader2025cloud}.
Recent Large Multimodal Models (LMMs) such as GPT-4V~\cite{hurst2024gpt}, LLaVA~\cite{liu2024llava}, and LISA~\cite{lai2024lisa,yang2023lisa++} demonstrate remarkable scene understanding for autonomous driving, including open-vocabulary recognition, reasoning-based segmentation, and natural-language situational awareness~\cite{li2025applications}. Deploying them in V2X architectures offers a promising path toward intelligent transportation systems that understand complex traffic beyond conventional perception pipelines~\cite{liu2025colmdriver}.

Despite this potential, naively streaming full camera frames from edge to cloud creates a dual bottleneck. In \textit{communication}, high-resolution driving images (e.g., $1600\times900$ in nuScenes~\cite{caesar2020nuscenes}) quickly saturate bandwidth-limited V2X links~\cite{jin2025bandwidth,huang2023v2x}; in \textit{computation}, LMMs convert images into long visual-token sequences whose transformer prefill scales \textit{quadratically} with input length, so full-frame transmission is inefficient for both network usage and cloud-side latency~\cite{shang2025llava,bolya2022token}.
Our key observation is spatial--temporal sparsity: \textit{safety-critical dynamic objects} (vehicles, pedestrians, cyclists) occupy only a small image region while most pixels are slowly varying background, so isolating and transmitting only Regions of Interest (ROIs) can reduce payload size and LMM token load without materially degrading perception~\cite{li2025mminference}.

Existing ROI selection strategies are mainly detection- or motion-based, and both leave a gap. \textit{Detection-based methods} (e.g., YOLO, RT-DETR~\cite{zhao2024rtdetr}) are efficient but output coarse boxes with substantial background, limiting compression, and rely on closed-set labels that are less robust to novel or context-dependent traffic entities~\cite{hatami2025openworld,yang2025openad}. \textit{Motion-based methods} (frame differencing or optical flow) are category-agnostic but easily corrupted by ego-motion, often confusing irrelevant motion (trees, shadows) with safety-critical actors and producing high false-positive ROI coverage~\cite{villar2024mcds,huang2025ego}.

\begin{figure*}[t]
	\centering
	\includegraphics[width=0.9\linewidth]{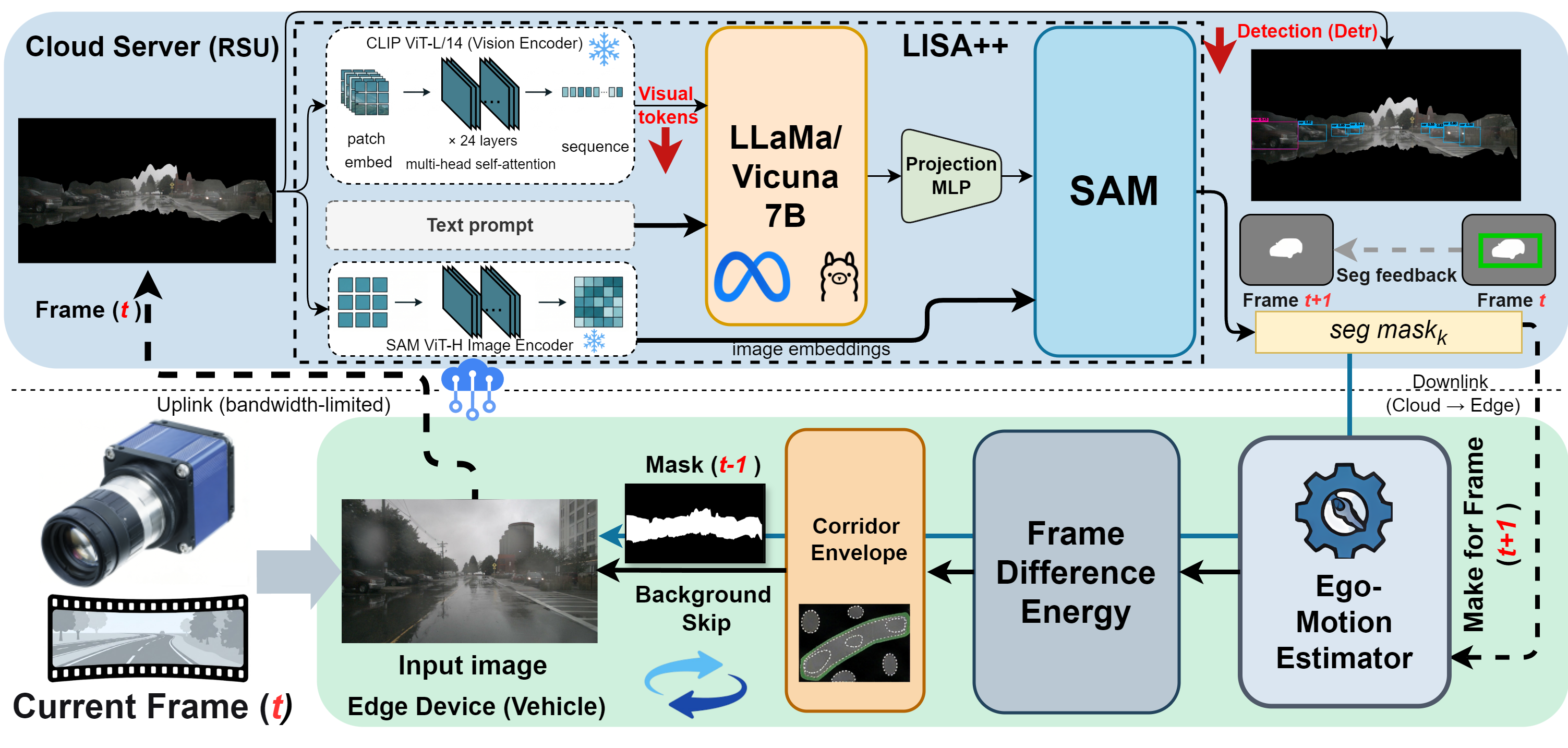}
	\caption{Overview of the proposed CABLE framework, a feedback-driven edge--cloud perception pipeline for bandwidth-efficient autonomous driving. The cloud server uses an LMM-based segmentation module to extract safety-critical object masks from the current frame, while the edge device reuses the returned mask together with ego-motion estimation and frame-difference energy to construct a corridor envelope for the next frame. This feedback design enables ROI-focused transmission and perception, reducing redundant background communication and cloud-side visual token processing.}
	\label{fig:overview}
\end{figure*}

To address these limitations, we propose \textbf{CABLE} (\textbf{C}loud-\textbf{A}ssisted \textbf{B}andwidth-efficient \textbf{L}MM-based \textbf{E}ncoding), an edge-cloud perception framework with cross-frame feedback for bandwidth-constrained V2X, as illustrated in Fig.~\ref{fig:overview}.
At each frame, the edge predicts the ROI using only lightweight operations (ego-motion compensation, frame-difference energy, ROI fusion, and a corridor envelope that links discrete regions into a continuous band), then uploads an ROI-only image with background zeroed.
The cloud runs LISA++~\cite{yang2023lisa++} on this compact ROI image to produce semantic masks with reduced visual tokens and lower inference latency.
The predicted mask is sent back to the edge as the prior for the next frame, forming a perception feedback loop: mask $\rightarrow$ ROI $\rightarrow$ LMM $\rightarrow$ mask. We use ``feedback'' to denote cross-frame information propagation in a tracking-by-segmentation sense, not closed-loop control in the control-theoretic sense: there is no reference signal or error term minimized over time.

The main contributions of this work are as follows:
\begin{enumerate}
	\item We propose CABLE, a cloud--edge perception framework that couples cloud-side LMM segmentation with edge-side ego-motion mask propagation, residual-motion refinement, and corridor envelope fusion, forming a cross-frame feedback loop where the cloud segmentation mask drives the next-frame ROI generation on the edge.
	\item We introduce a lightweight refresh-control rule using ROI-coverage and mask-consistency triggers to switch between ROI-only upload and keyframe refresh, improving robustness under cold start and drift.
	\item We design an adaptive background-skip scheme that transmits only the ROI-masked image, jointly reducing uplink bandwidth, LMM input tokens, and prefill latency while preserving fidelity for safety-critical actors.
	\item We validate CABLE on five benchmarks (nuScenes~\cite{caesar2020nuscenes}, WOD-ZB, Waymo~\cite{sun2020waymo}, KITTI~\cite{geiger2012kitti}, CADC~\cite{pitropov2021cadc}), achieving $73$--$87\%$ ROI pixel-coverage reduction and $5$--$8\times$ estimated prefill speedup while largely preserving detection performance (modest, bounded trade-off) across diverse geographies and weather.
\end{enumerate}

\section{Related Work}
\label{sec:related}

\subsection{Cooperative Perception in V2X Systems}
V2X cooperative perception overcomes single-vehicle limitations (e.g., occlusion) by sharing sensory data. While early methods transmit costly raw data~\cite{chen2019cooper}, intermediate fusion reduces bandwidth by sharing learned features. For instance, V2VNet~\cite{wang2020v2vnet} and DiscoNet~\cite{li2021learning} employ graph-based feature aggregation, while When2com~\cite{liu2020when2com}, Who2com~\cite{liu2020who2com}, and Where2comm~\cite{hu2022where2comm} optimize communication scheduling and spatial confidence to minimize bandwidth. A complementary line of work reduces communication by selecting only the most informative content to share, e.g., infrastructure-side critical feature extraction~\cite{zhang2024infracritical}, supply--demand-driven information selection~\cite{liu2024supplydemand}, and lifetime-aware feature collection for target coverage in heterogeneous visual sensor networks~\cite{wang2023lifetime}. Recent advances further explore vision transformers~\cite{xu2023v2xvit}, adversarial robustness~\cite{li2023colla}, and heterogeneous sensors~\cite{han2023collaborative}, evaluated on benchmarks like OPV2V~\cite{xu2022opv2v}, V2X-Sim~\cite{li2022v2xsim}, and DAIR-V2X~\cite{yu2022dair}. These methods assume edge devices can extract local features; CABLE instead addresses the unexplored case where a cloud-hosted LMM is the sole perception backbone, optimizing transmission purely via semantic ROI selection.

\subsection{Bandwidth-Efficient Edge--Cloud Inference}
Edge-cloud DNN deployment faces severe communication bottlenecks. Early split computing~\cite{kang2017neurosurgeon} partitioned layers based on latency constraints, while subsequent methods compress intermediate features at the split point using autoencoders~\cite{shao2020bottlenet}, joint optimization~\cite{eshratifar2019jointdnn}, knowledge distillation~\cite{matsubara2022bottlefit}, and saliency-guided bottlenecks~\cite{furutanpey2023frankensplit}. System-level solutions also explore dynamic split points~\cite{li2019edge} and edge architectures for autonomous driving~\cite{liu2019edge}, while image-level approaches optimize raw transmission via ROI-aware encoding~\cite{wang2024roiaware}, task-driven semantic coding~\cite{zhang2023taskdriven}, or tiled processing~\cite{chen2023tiledvlm}, with related constraints studied in federated learning~\cite{shi2020communication}. Compared with these (and ROI-aware encoding~\cite{wang2024roiaware} in particular), CABLE differs in that it (i)~acts in the raw-image domain rather than inside the codec/feature pipeline, requiring no model internals and supporting closed-source cloud LMMs; (ii)~drives its ROI from open-vocabulary semantic masks rather than a detector- or saliency-defined box; (iii)~feeds the cloud mask back as the next-frame prior instead of a one-shot or fixed ROI; and (iv)~jointly targets uplink bandwidth and LMM visual-token/prefill cost rather than bandwidth alone.

\subsection{Large Multimodal Models for Visual Perception}
LMMs merge visual encoders with language models for open-vocabulary understanding. LLaVA~\cite{liu2024llava} and LLaVA-1.5~\cite{liu2024llavav15} pioneered visual instruction tuning, while promptable foundation models such as SAM~\cite{kirillov2023sam} and SAM2~\cite{ravi2024sam2} advanced zero-shot segmentation and tracking. Bridging these, LISA~\cite{lai2024lisa} and LISA++~\cite{yang2023lisa++} introduced reasoning segmentation, generating multi-target masks from implicit text queries without category-specific training. As their quadratic prefill cost restricts such 7B-scale models to the cloud, CABLE harnesses LISA++~\cite{yang2023lisa++} while transmitting only the tracked ROI, and pairs it with RT-DETR~\cite{zhao2024rtdetr} for high-speed downstream detection; each cloud inference yields both perception results and the prior for the next frame's transmission.


\section{Proposed Method}
\label{sec:method}
CABLE operates a continuous feedback loop between a lightweight edge device and a cloud-hosted large multimodal model (LMM): the cloud segmentation mask from frame $k$ is propagated on the edge to generate the ROI for frame $k{+}1$, and only the ROI-masked image is transmitted to the cloud for the next segmentation. This section describes each component of the pipeline.

\subsection{Problem Formulation}

Consider a monocular camera stream $\{I_k\}_{k=0}^{N}$ at an edge device (vehicle or roadside unit) that has ego-motion estimates (velocity $v_x, v_y$ and yaw rate) but cannot run an LMM locally; a cloud server hosts an LMM $\mathcal{F}$ for open-vocabulary segmentation over a limited uplink. For each frame $I_k$, the edge constructs a binary ROI mask $R_k \in \{0,1\}^{H \times W}$ and transmits only the masked image $\tilde{I}_k = I_k \odot R_k$ (background zeroed); the cloud runs $\mathcal{F}$ on $\tilde{I}_k$ and returns a segmentation mask $M_k$ used as the prior for the next frame. Formally, CABLE seeks a frame-wise policy that minimizes communication load while keeping semantic coverage above a task-required level:
\begin{equation}
	\min_{\{R_k\}} \sum_{k=1}^{N} \frac{|R_k|}{H\!\cdot\!W}
	\quad
	\text{s.t.}
	\quad
	\mathcal{Q}(M_k, G_k) \ge \gamma,
\end{equation}
where $|R_k|/(H\!\cdot\!W)$ is ROI coverage ratio, $G_k$ denotes ground-truth object regions (available only for evaluation), $\mathcal{Q}(\cdot)$ is a perception-quality metric (e.g., recognition rate / detection retention), and $\gamma$ is a minimum acceptable quality threshold.

\subsection{Cloud Keyframe Initialization and Refresh}

For the first frame $I_0$ of each sequence, no prior mask is available. The full image is transmitted to the cloud, where the LMM performs open-vocabulary segmentation:
\begin{equation}
	M_0 = \mathcal{F}(I_0, \, p),
\end{equation}
where $p$ is a text prompt specifying safety-critical road actors (vehicles, pedestrians, cyclists), and $M_0$ serves as the initial prior. During normal operation CABLE transmits ROI-only images; when the propagated mask confidence drops (severe drift or near-empty valid ROI) the edge triggers a full-frame refresh via a lightweight trigger $c_k \in \{0,1\}$:
\begin{equation}
	c_{k+1} = \mathbf{1}\!\left[\frac{|R_k|}{H\!\cdot\!W} < \tau_{\mathrm{min}}\; \lor\; \frac{|\hat{M}_{k-1} \cap M_k|}{|\hat{M}_{k-1} \cup M_k|} < \tau_{\mathrm{iou}}\right],
\end{equation}
where $\tau_{\mathrm{min}}$ avoids degenerate near-empty ROI and $\tau_{\mathrm{iou}}$ detects mask inconsistency. If $c_{k+1}=1$, the next upload uses a full frame; otherwise, CABLE continues ROI-only transmission.

\subsection{Edge-Side Geometric Mask Propagation}

For subsequent frames $k > 0$, the edge propagates the previous cloud mask $M_{k-1}$ to the current viewpoint using ego-motion compensation.

\subsubsection{Ego-Motion Homography}
Given the ego velocity $(v_x, v_y)$ and camera intrinsics $(f_x, f_y)$, we approximate the inter-frame geometric transformation as a planar homography under a pure-translation assumption:
\begin{equation}
	\mathbf{H}_k = \begin{bmatrix} 1 & 0 & -\frac{f_x \cdot v_y \cdot \Delta t}{\max(|v_x \cdot \Delta t|,\, \epsilon)} \\ 0 & 1 & 0 \\ 0 & 0 & 1 \end{bmatrix},
\end{equation}
where $\Delta t = 1/f_{\mathrm{fps}}$ is the inter-frame interval and $\epsilon$ is a small constant to prevent division by zero. The warped mask and warped previous frame are obtained via perspective transformation:
\begin{equation}
	\hat{M}_{k-1} = \mathcal{W}(M_{k-1},\, \mathbf{H}_k), \quad \hat{I}_{k-1} = \mathcal{W}(I_{k-1},\, \mathbf{H}_k).
\end{equation}
This step provides a motion-aligned semantic prior before appearance-based refinement. The pure-translation approximation is justified by the short inter-frame interval $\Delta t$, over which the yaw increment is small and apparent motion is translation-dominated. CABLE does not require it to be exact: residual rotational misalignment during turns, lane changes, or intersections is absorbed by the residual-motion energy term, the yaw-aware buffer dilation $\beta|\Delta\psi_k|$, and the corridor envelope, while persistent drift triggers a full-frame keyframe refresh.

\subsubsection{Residual-Motion Energy Detection}
The geometric warp handles ego-motion but not independently moving objects entering the field of view; we detect such changes via a frame-difference energy map:
\begin{equation}
	E_k = \left| \, \mathrm{gray}(I_k) - \mathrm{gray}(\hat{I}_{k-1}) \, \right|, \quad
	D_k = \mathbf{1}[E_k > \tau_e],
\end{equation}
where $\tau_e$ is an energy threshold, so $D_k$ captures significant appearance change after ego-motion compensation (newly appeared or fast-moving objects). We use a fixed $\tau_e$ on the $[0,255]$ grayscale difference (value in Sec.~\ref{sec:setup}); since $D_k$ is later gated by the dilated mask buffer, a single global value generalizes across all five datasets, and an adaptive noise-driven variant is left to future work.

\subsubsection{Adaptive Buffer Dilation}
To prevent the propagated mask from shrinking over time due to accumulated warp errors, we dilate it with an ego-adaptive buffer:
\begin{equation}
	\hat{M}_{k-1}^{+} = \mathrm{dilate}(\hat{M}_{k-1},\, b_k), \quad
	b_k = b_0 + \alpha \|\Delta \mathbf{d}_k\| + \beta |\Delta \psi_k|,
\end{equation}
where $b_0$ is a base size, $\|\Delta \mathbf{d}_k\|$ and $|\Delta \psi_k|$ are inter-frame translation and yaw displacement, and $\alpha, \beta$ are gains, so the buffer grows with speed to prevent erosion during fast driving.

\subsubsection{ROI Fusion}
The raw ROI combines the warped mask with the motion-gated energy residual:
\begin{equation}
	R_k^{\mathrm{raw}} = \hat{M}_{k-1} \;\cup\; \bigl(\hat{M}_{k-1}^{+} \;\cap\; D_k\bigr).
\end{equation}
Gating $D_k$ by the dilated buffer $\hat{M}_{k-1}^{+}$ suppresses false triggers from global illumination changes or sensor noise while still capturing genuinely new objects near existing detections.

\subsection{Corridor Envelope Consolidation}

The fused ROI $R_k^{\mathrm{raw}}$ may consist of disconnected blobs, and transmitting scattered ROI patches fragments the spatial context available to the LMM. We thus consolidate the ROI into a contiguous corridor envelope that preserves inter-object spatial relationships. For each column $x$ in the active region, we extract the topmost and bottommost ROI pixels as boundary curves $\mathbf{t}(x)$ and $\mathbf{b}(x)$, smoothed via a moving-average filter of window $w_s$ and extended by fractional margins $\eta_t, \eta_b$:
\begin{equation}
	\begin{aligned}
		\mathbf{t}'(x) & = \mathrm{smooth}(\mathbf{t}(x),\, w_s) - \eta_t \cdot \mathbf{t}(x),       \\
		\mathbf{b}'(x) & = \mathrm{smooth}(\mathbf{b}(x),\, w_s) + \eta_b \cdot (H - \mathbf{b}(x)).
	\end{aligned}
\end{equation}
A minimum corridor height $h_{\min}$ is enforced to guarantee sufficient vertical extent. The final ROI mask $R_k$ is the filled polygon bounded by $\mathbf{t}'(x)$ and $\mathbf{b}'(x)$, unioned with $R_k^{\mathrm{raw}}$. Numerical values for $w_s$, $\eta_t$, $\eta_b$, and $h_{\min}$ are reported in Sec.~\ref{sec:setup}.

\subsection{Background-Skip Transmission and Cloud Inference}

The edge constructs a background-skipped image by zeroing all pixels outside the ROI:
\begin{equation}
	\tilde{I}_k(u,v) = \begin{cases} I_k(u,v) & \text{if } R_k(u,v) = 1 \\ \mathbf{0} & \text{otherwise} \end{cases}
\end{equation}
and transmits $\tilde{I}_k$ to the cloud. Since the ROI typically covers only $10$--$25\%$ of the image area, this yields a $73$--$87\%$ reduction in transmitted pixels. We report this ROI pixel-coverage reduction as a proxy for bandwidth savings; the actual on-the-wire bitrate depends on the entropy coder applied to the sparse image and is discussed as a limitation rather than measured on a live link.

On the cloud, the LMM segments $\tilde{I}_k$ with the same prompt $p$:
\begin{equation}
	M_k = \mathcal{F}(\tilde{I}_k, \, p) \;\cap\; R_k.
\end{equation}
The intersection with $R_k$ keeps the mask within the transmitted region. Assuming image tokens scale linearly with the ROI ratio $\rho_k = |R_k| / (H \!\cdot\! W)$, the input token count becomes $n_{\mathrm{in}} = n_p + \lfloor n_v \cdot \rho_k \rfloor$ ($n_p, n_v$ the prompt and full-image token counts), so the quadratic prefill cost is reduced by a factor of approximately $(n_{\mathrm{in}} / (n_p + n_v))^{-2}$.

\subsection{Cross-Frame Mask Feedback}

The cloud mask $M_k$ is transmitted back to the edge (a lightweight binary mask requiring minimal downlink bandwidth) and replaces the previous mask state:
\begin{equation}
	M_{k-1} \leftarrow M_k.
\end{equation}
This completes the feedback loop: the cloud segmentation continuously refines the edge's scene prior, which governs the next-frame ROI. Unlike open-loop approaches relying on a one-time detection, this lets the system adapt to evolving scenes, recover from transient segmentation failures, and track objects across frames without a dedicated edge tracker. Algorithm~\ref{alg:cable_online} summarizes the per-frame online pipeline.

\begin{algorithm}[t]
	\caption{CABLE Online Feedback Inference}
	\label{alg:cable_online}
	\begin{algorithmic}[1]
		\Require Stream $\{I_k\}_{k=0}^{N}$, prompt $p$, thresholds $\tau_{\mathrm{min}}, \tau_e, \tau_{\mathrm{iou}}$
		\Ensure ROI-masked uploads $\tilde{I}_k$ and cloud masks $M_k$
		\State Upload full frame $I_0$ to cloud and get $M_0 \gets \mathcal{F}(I_0, p)$
		\State Initialize refresh flag $c_1 \gets 0$
		\For{$k = 1$ to $N$}
		\State Warp prior mask: $\hat{M}_{k-1} \gets \mathcal{W}(M_{k-1}, \mathbf{H}_k)$
		\State Warp previous frame: $\hat{I}_{k-1} \gets \mathcal{W}(I_{k-1}, \mathbf{H}_k)$
		\State Compute residual motion: $D_k \gets \mathbf{1}[|\mathrm{gray}(I_k)-\mathrm{gray}(\hat{I}_{k-1})| > \tau_e]$
		\State Compute adaptive dilation: $\hat{M}_{k-1}^{+} \gets \mathrm{dilate}(\hat{M}_{k-1}, b_k)$
		\State Build raw ROI: $R_k^{\mathrm{raw}} \gets \hat{M}_{k-1} \cup (\hat{M}_{k-1}^{+} \cap D_k)$
		\State Corridor consolidation: $R_k \gets \mathrm{Corridor}(R_k^{\mathrm{raw}})$
		\If{$c_k = 1$}
		\State Upload full frame $I_k$ and run $M_k \gets \mathcal{F}(I_k, p)$
		\Else
		\State Upload ROI image $\tilde{I}_k \gets I_k \odot R_k$ and run $M_k \gets \mathcal{F}(\tilde{I}_k, p) \cap R_k$
		\EndIf
		\State Update next-frame refresh flag:
		\Statex \hspace{1.4em}$c_{k+1} \gets \mathbf{1}[|R_k|/(H\!\cdot\!W) < \tau_{\mathrm{min}}\ \lor\ \mathrm{IoU}(\hat{M}_{k-1}, M_k) < \tau_{\mathrm{iou}}]$
		\State Feedback update: $M_{k-1} \gets M_k$
		\EndFor
	\end{algorithmic}
\end{algorithm}

\section{Experiments}
\label{sec:experiments}
\begin{table*}[t]
	\centering
	\caption{Cross-dataset evaluation of CABLE (feedback + corridor envelope). Left: communication efficiency metrics. Right: RT-DETR detection quality comparing full-frame vs.\ ROI-only inference against dataset ground truth. Higher is better for all metrics except ROI Coverage.}
	\label{tab:cross_dataset}
	\resizebox{\textwidth}{!}{%
		\begin{tabular}{l cccc c | ccc ccc}
			\toprule
			                 & \multicolumn{4}{c}{\textbf{Communication Efficiency}} &                  & \multicolumn{3}{c}{\textbf{Full-Frame DETR}} & \multicolumn{3}{c}{\textbf{ROI-Only DETR}}                                                                   \\
			\cmidrule(lr){2-5} \cmidrule(lr){7-9} \cmidrule(l){10-12}
			\textbf{Dataset} & \textbf{ROI Cov.}                                     & \textbf{BW Save} & \textbf{Token Red.}                          & \textbf{Prefill}                           & \textbf{Recog.} & Recall & Prec. & F1   & Recall & Prec. & F1   \\
			                 & (\%)                                                  & (\%)             & (\%)                                         & ($\times$)                                 & (\%)            &        &       &      &        &       &      \\
			\midrule
			nuScenes         & 23.7                                                  & 76.3             & 55.8                                         & 5.2$\times$                                & 98.4            & 0.88   & 0.88  & 0.86 & 0.88   & 0.88  & 0.86 \\
			WOD-ZB           & 13.9                                                  & 86.1             & 62.9                                         & 7.5$\times$                                & 88.1            & 0.68   & 0.16  & 0.15 & 0.61   & 0.18  & 0.16 \\
			Waymo            & 20.3                                                  & 79.7             & 58.3                                         & 6.2$\times$                                & 80.9            & 0.45   & 0.81  & 0.54 & 0.32   & 0.77  & 0.40 \\
			KITTI            & 26.4                                                  & 73.6             & 53.8                                         & 5.3$\times$                                & 88.6            & 0.81   & 0.65  & 0.66 & 0.70   & 0.64  & 0.59 \\
			CADC             & 19.8                                                  & 80.2             & 58.6                                         & 6.0$\times$                                & 85.1            & 0.16   & 0.38  & 0.14 & 0.15   & 0.42  & 0.13 \\
			\midrule
			\textbf{Average} & 20.8                                                  & 79.2             & 57.9                                         & 6.0$\times$                                & 88.2            & 0.60   & 0.58  & 0.47 & 0.53   & 0.58  & 0.43 \\
			\bottomrule
		\end{tabular}%
	}
\end{table*}

\begin{figure}[thpb]
	\centering
	\includegraphics[width=\columnwidth]{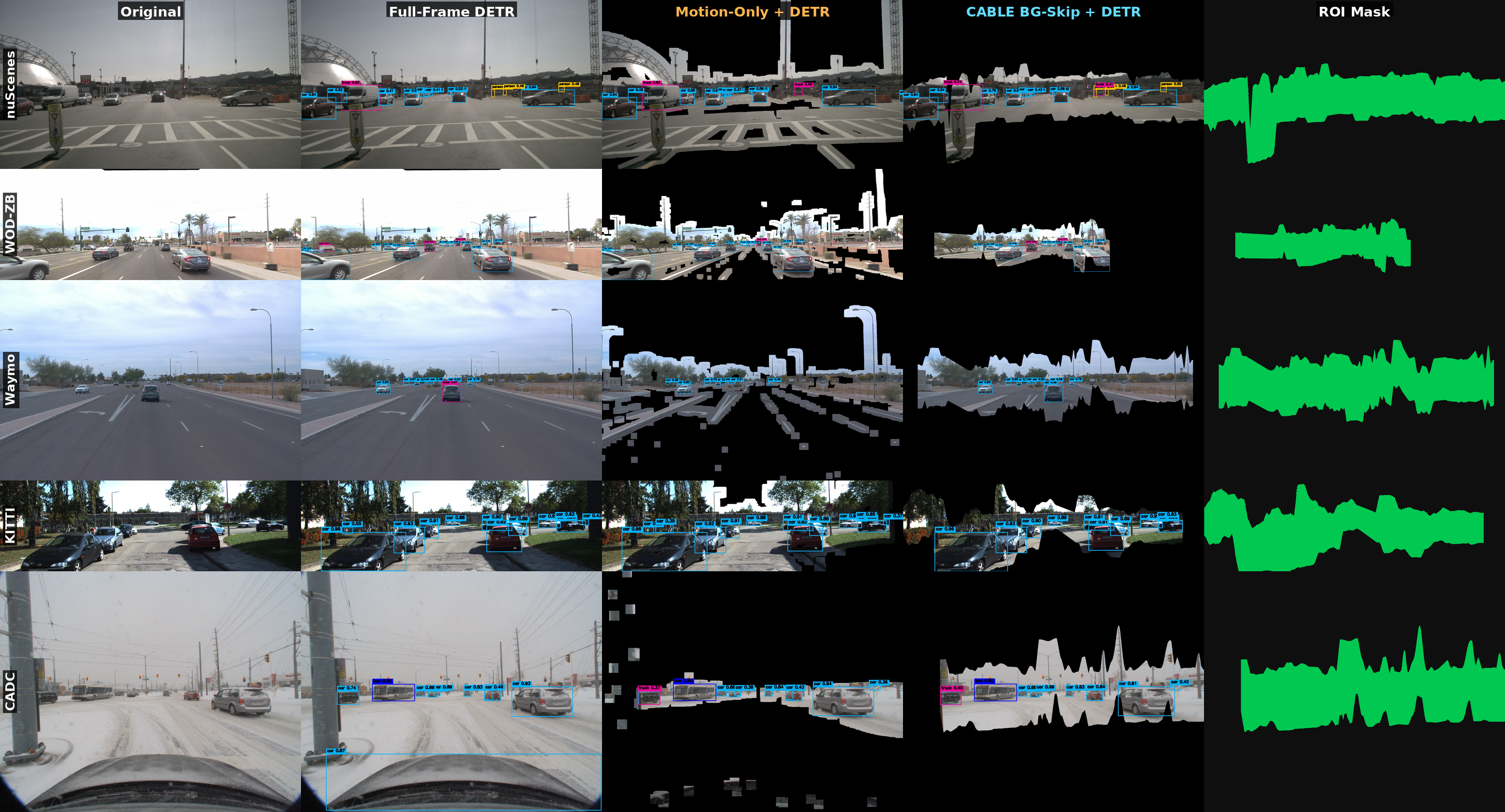}
	\caption{Qualitative comparison across five datasets. Each row corresponds to one dataset (nuScenes, WOD-ZB, Waymo, KITTI, CADC). From left to right: (a)~original frame, (b)~full-frame RT-DETR detections (baseline), (c)~motion-only background skip with RT-DETR (ROI derived purely from frame differencing, which retains excessive background and introduces noise), (d)~CABLE background skip with RT-DETR (ROI generated by LISA++ with corridor envelope, preserving a compact and spatially coherent driving corridor), and (e)~the CABLE ROI mask (green indicates transmitted pixels). Compared to motion-only ROI, CABLE achieves substantially smaller ROI regions while retaining all safety-critical objects, demonstrating consistent cross-dataset generalization under varying resolutions, aspect ratios, and weather conditions.}
	\label{fig:5dataset_comparison}
\end{figure}
\subsection{Experimental Setup}
\label{sec:setup}

\subsubsection{Datasets}
We evaluate CABLE on five datasets spanning diverse driving scenarios and weather conditions, using front-camera streams unless noted. \textbf{nuScenes}~\cite{caesar2020nuscenes} (10\,Hz) provides ground truth by projecting 3D vehicle/pedestrian boxes onto the image plane. \textbf{WOD-ZB} is a panoramic benchmark derived from WOD-E2E-R1 (built on WOD-E2E with SAM3-assisted masks/boxes and human-in-the-loop semantic labels), stitched into surround-view frames with dense COCO-format annotations at 10\,Hz. \textbf{Waymo}~\cite{sun2020waymo} and \textbf{KITTI Tracking}~\cite{geiger2012kitti} supply ego-motion from vehicle pose differences and oxts records, respectively. \textbf{CADC}~\cite{pitropov2021cadc} captures Canadian winter driving (snow, adverse weather) at $1280{\times}1024$, 3.3\,Hz, with ground truth from projected 3D cuboids (Car, Truck, Bus) via LiDAR--camera extrinsics.
\begin{table}[tpb]
	\centering
	\caption{ROI strategy comparison and corridor envelope ablation on nuScenes. All methods in (a) use corridor envelope post-processing. ``Det.\ Ret.'' is the fraction of full-frame RT-DETR detections preserved in ROI-only inference.}
	\label{tab:comparison_ablation}
	\setlength{\tabcolsep}{3pt}\resizebox{\columnwidth}{!}{%
		\begin{tabular}{ll cccccc}
			\toprule
			\textbf{Group} & \textbf{Method / Variant}    & \textbf{ROI Cov.}~(\%) & \textbf{BW Save}~(\%) & \textbf{Token Red.}~(\%) & \textbf{Prefill} & \textbf{Recog.}~(\%) & \textbf{Det.\ Ret.}~(\%) \\
			\midrule
			(a) Methods    & Motion-Only                  & 88.2                   & 11.8                  & 8.6                      & 1.4$\times$      & 98.0                 & 96.6                     \\
			               & Open-Loop LLM                & 56.6                   & 43.4                  & 31.8                     & 2.2$\times$      & 97.9                 & 96.7                     \\
			               & \textbf{CABLE (Ours)}        & 23.7                   & 76.3                  & 55.8                     & 5.2$\times$      & \textbf{98.4}        & \textbf{97.1}            \\
			\midrule
			(b) Ablation   & w/o Corridor                 & 6.0                    & 94.0                  & 68.7                     & 10.4$\times$     & 90.6                 & 42.7                     \\
			               & \textbf{w/ Corridor (CABLE)} & 23.7                   & 76.3                  & 55.8                     & 5.2$\times$      & \textbf{98.4}        & \textbf{97.1}            \\
			\bottomrule
		\end{tabular}%
	}
\end{table}
\subsubsection{Implementation Details}
The cloud-side LMM is LISA++~\cite{yang2023lisa++}, instantiated as \texttt{Senqiao/LISA\_Plus\_7b} (7B parameters, LLaVA-1.5~\cite{liu2024llavav15} backbone with SAM~\cite{kirillov2023sam} mask decoder).
The edge-side processing (homography warping, energy masking, corridor envelope fitting, and background skipping) runs purely on NumPy/OpenCV with no GPU requirement.
For detection evaluation, we employ RT-DETR~\cite{zhao2024rtdetr}, using \texttt{PekingU/rtdetr\_r50vd} (ResNet-50 backbone) with score threshold 0.4, and evaluate traffic-relevant COCO categories (person, bicycle, car, motorcycle, bus, truck) on both full-frame and ROI-only images.
All experiments run on a single NVIDIA RTX Pro 6000 for cloud inference and an Intel Core Ultra 285K with 128\,GB RAM for edge processing.
Unless otherwise noted, the edge-side hyperparameters are fixed across all five datasets (no per-dataset tuning): residual-motion energy threshold $\tau_e=25$ on the $[0,255]$ grayscale scale; refresh thresholds $\tau_{\mathrm{min}}=0.02$ (minimum ROI coverage) and $\tau_{\mathrm{iou}}=0.3$ (mask-consistency IoU); corridor smoothing window $w_s=15$ columns, fractional top/bottom margins $\eta_t=0.10$ and $\eta_b=0.15$, and minimum corridor height $h_{\min}=0.1\,H$; and adaptive-dilation parameters $b_0=8$\,px, $\alpha=1.5$, and $\beta=20$.

\subsubsection{Evaluation Metrics}
We report the following metrics:

\noindent\textbf{Communication Efficiency.} \emph{ROI Coverage} (\%) is the fraction of image pixels retained in the transmitted ROI, and \emph{Bandwidth Saving} (\%) $=100-\text{ROI Coverage}$. \emph{Token Reduction} (\%) is the estimated reduction in LMM input tokens (assuming token count scales linearly with image area), and \emph{Prefill Speedup} ($\times$) is the estimated prefill acceleration (assuming quadratic attention cost with input length).

\noindent\textbf{Perception Fidelity:}
As LISA++ reasoning masks are costly to score against box-level ground truth and not directly comparable across heterogeneous annotation formats, we quantify the downstream impact of ROI-only transmission with RT-DETR (a fast, reproducible detector applied identically to full-frame and ROI-only images), while the LMM's open-vocabulary masks are shown qualitatively in Fig.~\ref{fig:5dataset_comparison}. \emph{Object Recognition Rate} is the fraction of ground-truth objects whose box overlaps the ROI by at least 5\%, a permissive \emph{coverage} indicator (an object counts if at least partially transmitted), complemented by the stricter metrics below and discussed in Sec.~\ref{sec:limitations}. \emph{DETR Recall/Precision/F1} are RT-DETR metrics on ROI-only images against ground truth, and \emph{Detection Retention} is the fraction of full-frame RT-DETR detections preserved under ROI-only inference.

\subsection{Main Results}

\subsubsection{Cross-Dataset Evaluation}

Table~\ref{tab:cross_dataset} presents CABLE's performance across all five annotated datasets, with qualitative results shown in Fig.~\ref{fig:5dataset_comparison}. The pipeline consistently achieves substantial bandwidth savings while maintaining high object recognition across diverse geographies and weather conditions.

On nuScenes, CABLE achieves 76.3\% bandwidth saving at 98.4\% recognition and an estimated 5.2$\times$ prefill speedup, with ROI-only DETR matching full-frame inference (Recall 0.88/0.88, F1 0.86/0.86), confirming that the corridor envelope preserves sufficient spatial context. On WOD-ZB, ROI coverage drops to 13.9\% (86.1\% saving), exploiting temporal coherence over wide field-of-view panoramic sequences; the low baseline precision (0.16) reflects the domain gap to COCO-trained RT-DETR, while ROI-only precision slightly improves (0.18) as background removal reduces false positives. The pipeline generalizes without retraining to Waymo (79.7\% saving, 80.9\% recognition) and KITTI (73.6\% saving, 88.6\% recognition; ROI-only recall 0.81$\to$0.70 at stable precision). On CADC, despite snow-covered roads, CABLE attains 80.2\% saving at 85.1\% recognition and 6.0$\times$ speedup, and background removal again raises ROI-only precision (0.38$\to$0.42) by suppressing snow-induced false positives.

ROI-only transmission is not loss-free: averaged over the five datasets, F1 drops from 0.47 to 0.43 (largest on Waymo, 0.54$\to$0.40, and KITTI, 0.66$\to$0.59), where aggressive background removal clips object context near the ROI boundary. CABLE thus trades a modest, bounded perception cost for substantial communication and prefill savings.

\subsection{Comparison with Baselines}

We compare CABLE against two alternative ROI strategies on nuScenes, all using the same corridor envelope for fairness: (1)~\textbf{Motion-Only}: ROI from frame-difference energy alone (no semantic guidance); (2)~\textbf{Open-Loop LLM}: LISA++ segments only the first frame, subsequent frames rely on ego-motion propagation without cloud feedback; (3)~\textbf{CABLE (Feedback)}: full pipeline with per-frame cloud feedback.
We further ablate the corridor envelope by running CABLE without it (raw disjoint ROI mask); Table~\ref{tab:comparison_ablation} reports both the method comparison and this ablation.

As visualized in Fig.~\ref{fig:5dataset_comparison}(c--d), Motion-Only attains near-perfect recognition (98.0\%) but negligible bandwidth saving (11.8\%), as frame differences cover most of the image; Open-Loop LLM saves 43.4\% at 97.9\% recognition but degrades over time without feedback. CABLE (feedback) attains the highest detection retention (97.1\%) with strong recognition (98.4\%) and far higher savings (76.3\%), showing the value of the cloud feedback loop.

\subsection{Ablation Study}
As shown in Table~\ref{tab:comparison_ablation}(b), without the corridor envelope the ROI shrinks to 6.0\% coverage (94.0\% bandwidth saving, 10.4$\times$ prefill speedup), but the disjoint regions lose spatial context and detection retention collapses to 42.7\% as the detector fails to interpret fragmented patches. The corridor restores retention to 97.1\% and recognition to 98.4\%, confirming that its gap-filling is critical for preserving spatial reasoning.

\subsection{Limitations}
\label{sec:limitations}

Despite strong cross-dataset results, CABLE has limitations.
\emph{(i)~Cold start/drift:} it relies on full-frame cloud inference to recover the prior, and poor initialization can destabilize early ROI estimation.
\emph{(ii)~Ego-motion model:} the pure-translation homography degrades under large rotation (turns, lane changes, intersections), mitigated but not eliminated by residual-motion energy, yaw-aware dilation, corridor consolidation, and keyframe refresh.
\emph{(iii)~Permissive threshold:} the 5\% overlap criterion may overstate coverage for safety-critical objects needing full context, only partly offset by Detection Retention/F1.
\emph{(iv)~Baselines:} we do not yet benchmark against edge detectors, edge segmentation, or codec-integrated ROI encoding.
\emph{(v)~Deployment:} bandwidth is ROI pixel-coverage and prefill an analytic estimate; post-codec bitrate, an end-to-end latency breakdown (edge, uplink, cloud, downlink) over a real link, and on-vehicle deployment remain future work.






\section{Conclusion}
\label{sec:conclusion}

This paper presents CABLE, an edge--cloud perception framework with cross-frame feedback for bandwidth-constrained V2X systems. By feeding cloud-side LMM masks back to the edge for next-frame ROI generation, CABLE forms a mask-to-ROI-to-LMM feedback loop, combining ego-motion propagation, residual-motion refinement, and corridor consolidation to jointly cut transmission and visual-token load. Across five datasets (nuScenes, WOD-ZB, Waymo, KITTI, CADC), it attains 73--87\% ROI pixel-coverage reduction and 5--8$\times$ estimated prefill speedup at a modest, bounded detection trade-off; ablations show feedback improves robustness over open-loop propagation and the corridor is essential for spatial context. Future work targets adaptive keyframe scheduling, uncertainty-aware ROI expansion, edge-detector and codec baselines, and end-to-end bitrate/latency evaluation over real wireless links.

\bibliographystyle{IEEEtran}
\bibliography{ref}

\end{document}